\documentclass[sts]{imsart}

\RequirePackage{amsthm,amsmath,amsfonts,amssymb,dsfont}
\RequirePackage[numbers]{natbib}

\startlocaldefs

\usepackage{hyperref,cleveref}
\Crefformat{equation}{#2(#1)#3}
\crefformat{equation}{#2(#1)#3}
\Crefrangeformat{equation}{#3(#1)#4--#5(#2)#6}
\crefrangeformat{equation}{#3(#1)#4--#5(#2)#6}
\Crefmultiformat{equation}{#2(#1)#3}{ and~#2(#1)#3}{, #2(#1)#3}{ and~#2(#1)#3}
\crefmultiformat{equation}{#2(#1)#3}{ and~#2(#1)#3}{, #2(#1)#3}{ and~#2(#1)#3}

\usepackage{csquotes}
\MakeOuterQuote{"}

\usepackage{mathtools}
\DeclarePairedDelimiterXPP\ind[1]{\mathds{1}}{\lbrace}{\rbrace}{}{#1}
\DeclarePairedDelimiterX\eval[1]{\lbrace}{\rvert}{#1 \delimsize\rbrace}
\DeclarePairedDelimiter\ip{\langle}{\rangle}

\DeclarePairedDelimiter\norm{\lVert}{\rVert}

\DeclarePairedDelimiter\del{\lparen}{\rparen}
\DeclarePairedDelimiter\sbr{\lbrack}{\rbrack}

\DeclarePairedDelimiter\set{\lbrace}{\rbrace}

\DeclarePairedDelimiter\intcc{\lbrack}{\rbrack}

\def\ddefloop#1{\ifx\ddefloop#1\else\ddef{#1}\expandafter\ddefloop\fi}
\def\ddef#1{\expandafter\def\csname bf#1\endcsname{\ensuremath{\mathbf{#1}}}}
\ddefloop abcdefghijklmnopqrstuvwxyzABCDEFGHIJKLMNOPQRSTUVWXYZ\ddefloop
\def\ddef#1{\expandafter\def\csname bf#1\endcsname{\ensuremath{\boldsymbol{\csname #1\endcsname}}}}
\ddefloop {alpha}{beta}{gamma}{delta}{epsilon}{varepsilon}{zeta}{eta}{theta}{vartheta}{iota}{kappa}{lambda}{mu}{nu}{xi}{pi}{varpi}{rho}{varrho}{sigma}{varsigma}{tau}{upsilon}{phi}{varphi}{chi}{psi}{omega}{Gamma}{Delta}{Theta}{Lambda}{Xi}{Pi}{Sigma}{varSigma}{Upsilon}{Phi}{Psi}{Omega}{ell}\ddefloop
\def\ddef#1{\expandafter\def\csname cal#1\endcsname{\ensuremath{\mathcal{#1}}}}
\ddefloop ABCDEFGHIJKLMNOPQRSTUVWXYZ\ddefloop

\newcommand\R{\ensuremath{\mathbb{R}}} %
\newcommand\E{\ensuremath{\mathsf{E}}} %
\newcommand\reg{\ensuremath{f}} %
\newcommand\link{\ensuremath{g}} %
\newcommand\fmap{\ensuremath{\Phi}} %
\newcommand\lmap{\ensuremath{L}} %
\newcommand\T{{\ensuremath{\scriptscriptstyle{\mathsf{T}}}}} %
\newcommand\ran{\ensuremath{\operatorname{ran}}} %
\newcommand\indep{\ensuremath{\perp\!\!\!\!\perp}} %
\newcommand\Proj{\ensuremath{\mathsf{P}}} %
\newcommand\var{\ensuremath{\operatorname{var}}} %
\newcommand\argmax{\ensuremath{\operatorname*{arg\,max}}} %
\newcommand\argmin{\ensuremath{\operatorname*{arg\,min}}} %
\newcommand\sign{\ensuremath{\operatorname{sign}}} %
\newcommand\width{\ensuremath{\mathsf{w}}} %
\newcommand\EGOP{\ensuremath{\Gamma}} %

\endlocaldefs

\begin{document}

\begin{frontmatter}
\title{Survey on algorithms for multi-index models}
\runtitle{Algorithms for multi-index models}

\begin{aug}
\author[A]{\fnms{Joan}~\snm{Bruna}\ead[label=e1]{bruna@cims.nyu.edu}}
\and
\author[B]{\fnms{Daniel}~\snm{Hsu}\ead[label=e2]{djhsu@cs.columbia.edu}}
\address[A]{Courant Institute of Mathematical Sciences and Center for Data Science, New York University\printead[presep={\ }]{e1}.}
\address[B]{Department of Computer Science and Data Science Institute, Columbia University\printead[presep={\ }]{e2}.}
\end{aug}

\begin{abstract}
We review the literature on algorithms for estimating the index space in a multi-index model.
The primary focus is on computationally efficient (polynomial-time) algorithms in Gaussian space, the assumptions under which consistency is guaranteed by these methods, and their sample complexity.
In many cases, a gap is observed between the sample complexity of the best known computationally efficient methods and the information-theoretical minimum.
We also review algorithms based on estimating the span of gradients using nonparametric methods, and algorithms based on fitting neural networks using gradient descent.
\end{abstract}


\end{frontmatter}

\section{Introduction}

In a \emph{multi-index model} for data $(\bfx, \bfy)$ from $\R^d \times \R$, the regression function $\reg(x) := \E\sbr{ \bfy \mid \bfx = x}$ is assumed to depend only on a (possibly low-rank) linear transformation of the input:
\begin{equation} \label{eq:model_mim}
  \reg(x) = \E\sbr{\bfy \mid \bfx = x} = \link\del{\lmap x}
\end{equation}
for some rank-$r$ matrix $\lmap \in \R^{r \times d}$ whose row space $\ran(\lmap^\T)$ is called the \emph{index space}, and some function $\link \colon \R^r \to \R$ called the \emph{link function}.
When the number of variables $d$ is large, it is common to assume that $r \ll d$, so that the linear map $x \mapsto \lmap x$ is viewed as a form of dimension reduction that captures the information sufficient to optimally predict the response $\bfy$ from $\bfx$ under a mean squared error criterion.\footnote{%
  A potentially broader aim is to identify a linear map $L$ such that $\bfy \indep \bfx \mid L\bfx$.
  Goals such as this (as well as estimation goals concerning the model in~\Cref{eq:model_mim}) are the subject of the subfield of statistics known as \emph{Sufficient Dimension Reduction}~\citep{li1991sliced}, from which many of the methods and ideas discussed in this article are derived.%
}
(We postpone issues of identifiability for now.)
The special case where $r=1$ is known as the \emph{single-index model}.

The multi-index model is a popular model with benign high-dimensional behavior, with a rich and long history in the statistics literature, eg \cite{https://doi.org/10.1111/j.2517-6161.1964.tb00553.x, 66ace186-d33e-34c6-b587-c776d756007c, 7dc1b2c4-49df-3797-9492-fc8d6b538b08, li1991sliced} and references therein. More recently, it has also gained interest as a model for studying the ability of machine learning methods to perform "feature learning", ie to automatically discover meaningful low-dimensional structure within high-dimensional data. The model in~\Cref{eq:model_mim} is a special case of the more general model for $(\bfx,\bfy)$ where
\begin{equation} \label{eq:model_fl}
  \reg(x) = \link\del{\fmap(x)}
\end{equation}
and $\fmap \colon \R^d \to \R^r$ is a (possibly non-linear, but morally simpler than the original $f$) \emph{feature map}.
Feature learning, then, refers to the estimation of the feature map $\fmap$ from data.
The feature map may serve an explanatory role in understanding a predictor of $\bfy$ from $\bfx$, in which case feature learning is an end in itself.
Additionally, separating the tasks of learning $\fmap$ and learning $\link$ may be methodologically preferable; in such a scenario, feature learning is the first part of a two (or more) stage learning process.

Many recent works, under the guise of "multi-task learning" or "meta-learning", instantiate the model in~\Cref{eq:model_fl} separately for each of multiple data sources, but constrain the feature map $\fmap$ to be shared across all models.
For instance, in a computer vision context, the data sources may correspond to different object detection tasks in images; the shared feature map may give a semantic representation of images that is broadly useful for vision.
The data for a single task may be too specialized and/or insufficient to identify/estimate such a feature map, but the pooling of data sources across a diverse collection of tasks may pin down the desired image representation $\fmap$.
It has been hypothesized that large neural networks trained on diverse data sets may be encode such feature maps in the intermediate layers of the network, and hence feature learning has been considered as a possible explanation for the success of neural networks in practice.
See, e.g., \cite{du2021few} for more discussion of this compelling motivation.

The goal of this article is to survey algorithms for estimating the index space $\ran(\lmap^\T)$ in the multi-index model~\Cref{eq:model_mim}.
We do not attempt to be exhaustive, nor do we attempt to present the strongest or most general possible methods or results.
Rather, we aim to highlight some key ideas and methods from the literature, and to draw connections to the recent literature on gradient-based training of neural networks and statistical-computational gaps in high-dimensional inference.

\paragraph*{Acknowledgements:} The authors would like to thank Alex Damian, Adam Klivans, Florent Krzakala, Jason Lee, Luis Rademacher, Denny Wu and the anonymous referees for useful feedback during the completion of this work.

\section{Preliminaries}

\subsection{Identifiability}
\label{sec:ident}

We first observe that the model (\ref{eq:model_mim}) needs an additional `minimality' condition to be well-defined; hereafter, we will view (\ref{eq:model_mim}) as the representation of $f$ with \emph{smallest} rank $r$, which will be denoted the \emph{intrinsic dimension} of $f$. 
We can assume without loss of generality that $L$ is orthogonal, i.e., $L^\T \in \mathrm{Stiefel}(d, r)$, the (compact) Stiefel manifold of $d \times r$ matrices with orthonormal columns. 

It will be also convenient to formally define a multi-index model in terms 
of its generative process: the joint distribution $\pi$ of $(\bfx, \bfy)$ can be factorised into `uninformative' and `informative' components $\pi(x,y) = \pi_0( L_\perp x) \mathsf{G}( L x, y)$, where $\mathsf{G} \in \mathcal{P}( \R^r \times \R)$ is a squared-integrable $(r+1)$-dimensional probability measure satisfying 
$\mathsf{G} \neq \mathsf{G}_z \otimes \mathsf{G}_y$, and where $\mathsf{G}_z$ and $\mathsf{G}_y$ denote respectively the marginal of $\mathsf{G}$ along the first $r$ variables and the last variable. The link function $g$ is thus the conditional expectation $g(z) = \mathbb{E}_{\mathsf{G}}[ \bfy | \bfz = z]$. 
For instance, a deterministic multi-index model is defined as $\mathsf{G}(z,y) = \mathsf{G}_z(z) \delta( y - g(z))$. 

We say a linear subspace $W \subseteq \R^d$ is a \emph{mean dimension-reduction subspace} for $(\bfx,\bfy)$ if $\bfy \indep \reg(\bfx) \mid \Proj_W \bfx$ where $\Proj_W$ is a linear projector for $W$.
(Recall that $\reg(x) := \E\sbr{ \bfy \mid \bfx = x}$ is the regression function.)
Let $W_\cap$ be the intersection of all mean dimension-reduction subspaces for $(\bfx,\bfy)$.
If $W_\cap$ itself is a mean dimension-reduction subspace, then we say that $W_\cap$ is the \emph{central mean subspace (CMS)} for $(\bfx,\bfy)$.
The existence of the CMS for $\bfy \mid \bfx$ is guaranteed under rather mild conditions on the support of $\bfx$~\cite{cook2002dimension}.
For example, the CMS exists if the support of $\bfx$ is open and convex.
It is therefore natural to consider the multi-index model~\eqref{eq:model_mim} under conditions that ensure the existence of the CMS, and then to let $\lmap$ be a matrix whose rows form an orthonormal basis for the CMS.
This setup ensures the identifiability of $W:=\ran(\lmap^\T)$.
We adopt this setup throughout the survey (possibly with additional assumptions that also imply the existence of the CMS).

We will be interested in this question from two different perspectives: first, by viewing the multi-index model as a goal on itself, we will study dedicated algorithms, culminating in `optimal' methods, in a sense that will be precised later. 
Next, by viewing multi-index models as a template for `feature learning', we will describe the behavior of a `canonical' high-dimensional learning algorithm, notably gradient-descent methods on simple neural networks, when fed data generated by a multi-index model.

\subsection{Information-Theoretic Estimation Limits}

Under general conditions, and assuming that the previous identifiability condition, one expects that the required number of samples to estimate $W = \text{ran}(L^\perp)$ up to error $\epsilon$ will be of order $n= O( d r / \epsilon^2)$, by standard covering arguments \cite{dudeja2024statisticalcomputational, damian2024computational}.
Indeed, this estimator is constructed by building an appropriate 
$\epsilon$-net over the manifold of $r$-dimensional subspaces $W_1, \ldots, W_N$ in $\R^d$, of dimension $\sim dr$,
and choosing the subspace of highest likelihood based on the observed data.
This estimator is however generally intractable, since it amounts to optimizing a non-convex objective. 
The natural question is therefore how to design efficient algorithms to estimate the subspace, and what is their required sample complexity. This will be the focus of the next sections.

\subsection{Notations}

We use bold face symbols (e.g., $\bfx$, $\bfz$) to denote random variables.
Let $\gamma$ denote the standard Gaussian distribution on the real line, and $\gamma_d$ denote the standard Gaussian distribution in $\R^d$.
We consider the Lebesgue space $L^2(\R^d,\gamma_d)$ and write inner products as $\ip{f,g}_{\gamma_d} = \E_{\bfz \sim \gamma_d}[ f(\bfz) g(\bfz) ]$.
Let $H_k \colon \R^d \to (\R^d)^{\otimes k}$ denote the order-$k$ (normalized probabilist's) Hermite tensor~\citep{mccullagh2018tensor}, given by
\begin{equation*}
  H_k(u) = \frac{(-1)^k}{\sqrt{k!}} \frac{\nabla^k \gamma_d(u)}{\gamma_d(u)} , \quad u \in \R^d .
\end{equation*}
When $d=1$, $H_k$ becomes the degree-$k$ (normalized probabilist's) Hermite polynomial; the normalization is in $L^2(\R,\gamma)$, ensuring $\E_{\bfz \sim \gamma}[ h_k(\bfz) h_l(\bfz) ] = \ind{k=l}$.
We use $\mathrm{Stiefel}(d,r)$ to denote the Stiefel Manifold of $d \times r$ orthogonal matrices, and $\mathrm{Grassman}(d,r)$ for the Grassmann Manifold of $r$-dimensional subspaces of $\R^d$.

\section{The Gaussian setting}
\label{sec:moments}

We will first describe a general framework to estimate the support that 
makes strong assumptions on the input data distribution $\pi$, but in exchange enables very general choices for the link function. 
In particular, the Gaussian setting where $\bfx \sim \gamma_d$ will take centerpiece, even if some of the methods described below extend to more general settings.\footnote{%
  The remarkable work of~\cite{chandrasekaran2024smoothed} uses a smoothed analysis framework to analyze algorithms for learning multi-index models under \emph{worst-case} distributions on $\bfx$.
  In their framework, the goal of estimation is to compete against smoothed target functions.
  Also, the works of \cite{klivans2024learning,chandrasekaran2024efficient} consider learning in the presence of distribution-shift, where the distribution of $\bfx$ may differ from Gaussian at ``test time''.%
}
As discussed in \Cref{sec:ident}, this guarantees the identifiability of $W := \ran(\lmap^\T)$.
For simplicity, we also assume $\link$ is smooth (though it will be clear that only weak derivatives are needed).
These assumptions enable fairly simple moment-based estimators of $W$ that are computationally tractable.

\subsection{Linear estimator}

We initiate our survey with arguably the simplest estimator for the support in the single-index case (where $r = \dim(W) = 1$).
The following linear estimator was studied by~\cite{brillinger1982generalized}:
the estimate of $W$
from an i.i.d.~sample $(\bfx_1,\bfy_1),\dotsc,(\bfx_n,\bfy_n)$
is the line spanned by
\begin{equation} \label{eq:est_linear}
  \hat\bfv := \frac1n \sum_{i=1}^n \bfy_i \bfx_i .
\end{equation}
(This estimator is related to the "Average Derivative Estimator" of \cite{hardle1989investigating}, as the analysis below will make clear.)
In this case, we may write $\lmap = u^\T$ for a unit vector $u \in S^{d-1} := \set{ v \in \R^d \colon \norm{v}_2 = 1 }$ that spans $W$.
By Stein's lemma~\citep{stein1981estimation} and the chain rule,
\begin{align*}
  \E\sbr{\hat\bfv}
  & = \E\sbr{\bfy \bfx} = \E\sbr{\reg(\bfx) \bfx} \\
  & = \E\sbr{\nabla \reg(\bfx)} = \E\sbr{\link'(u^\T\bfx)} u \\
  & = \E\sbr{\link'(\bfz)} u
\end{align*}
where $\bfz \sim \gamma$ is a standard normal random variable.
The mean of $\hat\bfv$ is therefore always in $W$.
Moreover, if the expected derivative of $\link$ is non-zero---i.e., $\E\sbr{\link'(\bfz)} \neq 0$---then the law of large numbers implies the consistency of the linear estimator for estimating $W$.
Indeed, using the rotational symmetry of the Gaussian measure, we can assume w.l.o.g. that $\bfy = g( \bfx_1)$, and thus $\E\sbr{\| \bfy \bfx \|^2} = O(1) + (d-1) \simeq d$, leading to 
\begin{align*}
    \frac{\E\sbr{\| \hat\bfv - \E\sbr{\hat\bfv}\|^2}}{\| \E\sbr{\hat\bfv}\|^2} &\simeq \frac{d}{n} ~,
\end{align*}
indicating that $n \gg d$ samples are sufficient to obtain an accurate estimate in this setting of $\E\sbr{\link'(\bfz)} \neq 0$. For a more precise statement we refer the reader to \cite[Lemma F.11]{damian2024computational}.

\subsection{Noisy one-bit compressed sensing}

\cite{plan2012robust} studied the linear estimator in the context of noisy one-bit compressed sensing, where the aim is to recover a signal vector $u$ using linear measurements $u^\T\bfx$ that are quantized, say, to values in $\set{-1,1}$.
A natural "noise-free" variant of this problem assumes $\bfy = \sign(u^\T\bfx)$.
More generally, the measurements may be corrupted by noise in a way such that $\E\sbr{\bfy \mid \bfx} = \link(u^\T\bfx)$ for some unknown function $\link \colon \R \to \intcc{-1,1}$, so there is some chance that $\bfy \neq \sign(u^\T\bfx)$.
For example, if the sign is flipped independently with a constant probability $\eta \in \intcc{0,1}$, then $\link(z) = (1-2\eta)\sign(z)$.
(This special case was studied by \cite{servedio1999pac} under the guise of PAC learning homogeneous half-spaces with random classification noise under spherically symmetric distributions.)
The requirement that $\E\sbr{\link'(\bfz)} \neq 0$ can be regarded as a minimum signal strength condition for the linear estimator to work.
For example, this is satisfied by strictly monotone link functions like $\link(z) = \tanh(z)$.
(Note that it is different from assuming $\Pr\sbr{\bfy \neq \sign(u^\T\bfx)} \neq 1/2$.)

In compressed sensing, it is often assumed that the signal $u$ comes from a structured set, say, $K \subseteq B^d$, where $B^d$ is the $d$-dimensional Euclidean unit ball.
For example, $K$ may be the set of sparse vectors, or vectors with low $\ell^1$-norm.
\cite{plan2012robust} reinterpret the linear estimator as being the line spanned by
\begin{equation} \label{eq:est_linear_K}
  \hat\bfu := \argmax_{u \in K} \ip*{ u, \frac1n \sum_{i=1}^n \bfy_i \bfx_i } ,
\end{equation}
with $K = B^d$; they propose using the structured set $K$ in \Cref{eq:est_linear_K} when one has the prior knowledge $u \in K$.
Under the assumption $\E\sbr{\link'(\bfz)} \neq 0$, they show that the sample size $m$ needed by this modified linear estimator to accurately estimate $u$ scales only with (the square of) the \emph{Gaussian mean width $\width(K)$ of $K$}, which can be much smaller than the dimension $d$.
For example, if $K = \set{ v \in \R^d \colon \norm{v}_2 \leq 1 ,\, \norm{v}_1 \leq \sqrt{s} }$, then $\width(K) = O(\sqrt{s\log(2d/s)})$.
Furthermore, if $K$ is convex, then optimization problem in \Cref{eq:est_linear_K} is a convex optimization problem, which can be solved efficiently under fairly mild conditions on $K$.

In summary, the assumption $\E\sbr{\link'(\bfz)} \neq 0$ enables efficient estimation methods with a rate matching the information-theoretic bound, and are thus `optimal' in this sense of sample complexity. 
The natural question is then to understand how to proceed when this property does not hold.

\subsection{Principal Hessian Directions}
\label{sec:hessian}
For the general multi-index case (where $r > 1$ is allowed), \cite{li1992principal} proposed the "Principal Hessian Directions" (PHD) estimator, which in the present setting with normal $\bfx$, is defined by the span of the eigenvectors corresponding to the $r$ largest (in magnitude) eigenvalues of
\begin{equation} \label{eq:est_phd}
  \widehat\bfM := \frac1n \sum_{i=1}^n \bfy_i \del{ \bfx_i \bfx_i^\T - I_d } .
\end{equation}
Using (the second-order version of) Stein's lemma and the chain rule again,
\begin{align*}
  \E\sbr{\widehat\bfM}
  & = \E\sbr{\bfy \del{\bfx\bfx^\T - I_d}} = \E\sbr{\reg(\bfx) \del{\bfx\bfx^\T - I_d}} \\
  & = \E\sbr{\nabla^2 \reg(\bfx)} = L^\T \E\sbr{\nabla^2 \link(\lmap^\T\bfx)} L \\
  & = L^\T \E\sbr{\nabla^2 \link(\bfz)} L
\end{align*}
where $\bfz \sim \gamma_r$ is now an $r$-dimensional standard normal random vector.
The range of the mean of $\widehat\bfM$ is therefore a subspace of $W$.
If, additionally, the expected Hessian $\E\sbr{\nabla^2 \link(\bfz)}$ is non-singular, then the range of $\E\sbr{\widehat\bfM}$ is, in fact, equal to $W$; we say that the estimator is \emph{exhaustive} in this case.
When PHD is exhaustive, the law of large numbers implies that it is a consistent estimator of $W$.
Again, one can verify \cite{mondelli2018fundamental, lu2020phase} that under this non-singular assumption (playing the analog of the moment assumption $\E\sbr{\link'(\bfz)} \neq 0$ in the linear case), and in the single-index setting, the required sample complexity to recover an accurate estimate is $n = \Theta(d)$, with a constant that depends on the link function.

\subsection{Application: Learning Convex Concepts}

A dimension-reduction technique of \cite{vempala2010learning} for PAC learning low-dimensional convex concepts turns out to be a special case of PHD.
Suppose $\reg(x) = \ind{x \in K}$ for some convex set $K \subseteq \R^d$ whose supporting hyperplanes' normal vectors span $W$.
In other words, determining membership of $x$ in $K$ is equivalent to checking membership of $\Proj_W x$ in $W$.
The goal is to learn a hypothesis $h \colon \R^d \to \set{0,1}$ with low error rate $\Pr[h(\bfx) \neq \reg(\bfx)]$.
If $K$ is a full-dimensional convex set, then $W = \R^d$, and the problem appears to be computationally intractable in general (see, e.g., \citep{klivans2008learning}).
However, if $\dim(W)$ is small, then dimension reduction can be used to reduce the computational difficulty.

Assume for simplicity that $K \cap W$ is a symmetric convex body in $W$ with positive probability mass.
First, since $\bfy$ is $\set{0,1}$-valued, we can write $\E\sbr{\widehat\bfM}$ as
\begin{equation*}
  \E\sbr{\widehat\bfM}
  = \E\sbr{\bfy \del{\bfx\bfx^\T - I_d}}
  = \E\sbr{ \bfy } \E\sbr{ \bfx\bfx^\T - I_d \mid \bfy = 1 } .
\end{equation*}
Furthermore, we have $\E\sbr{ \bfy } = \Pr\sbr{ \reg(\bfx) = 1 } \neq 0$ by assumption.
If $v \in W^\perp \cap S^{d-1}$, then
\begin{align*}
  v^\T \E\sbr{\widehat\bfM} v
  & = \E\sbr{\bfy} \del{ \E\sbr{ (v^\T\bfx)^2 \mid \bfy = 1 } - 1 } \\
  & = \E\sbr{\bfy} \del{ \E\sbr{ (v^\T\bfx)^2 } - 1 } = 0
\end{align*}
since $v^\T\bfx \indep \bfy$.
Moreover, for $v \in W \cap S^{d-1}$,
\begin{align*}
  v^\T \E\sbr{\widehat\bfM} v
  & = \E\sbr{\bfy} \del{ \E\sbr{ (v^\T\bfx)^2 \mid \bfy = 1 } - 1 } \\
  & = \E\sbr{\bfy} \del{ \var\del{ v^\T\bfx \mid \bfy = 1 } - 1 } < 0
\end{align*}
since the variance of a truncated standard normal distribution is strictly less than one;
see~\cite[Appendix~B and Appendix~C]{klivans2024learning} for quantitative bounds.
This shows that the range of $\E\sbr{\bfy \del{\bfx\bfx^\T - I_d}}$ is precisely $W$.
Projecting the data to the PHD subspace reduces the computational difficulty of learning convex concepts (say, using the general polynomial regression technique of \cite{klivans2008learning}), because one now only has to work in $r$-dimensional space as opposed to the original $d$-dimensional space.

\subsection{The Information Exponent for Single-Index Models}
\label{sec:infoexpo}

An important limitation of the linear and PHD estimators is that they may be non-exhaustive depending on the link function.
For example, $\link(z) = z^2$ (as considered in the "phase retrieval" problem) has $\E\sbr{\link'(\bfz)} = \E\sbr{2\bfz} = 0$, so the linear estimator is non-exhaustive in this case.
And $\link(z) = z^3 - 3z$ has $\E\sbr{\link'(\bfz)} = \E\sbr{3\bfz^2 - 3} = 0$ and $\E\sbr{\nabla^2 \link(\bfz)} = \E\sbr{ 6\bfz } = 0$, so both the linear and PHD estimators are non-exhaustive in this case.

At this point, it is apparent that there should be a general 
principle at play that relates a generic estimation procedure with a structural property of the link function, the number of vanishing moments of the form $\E\sbr{ \bfy p(\bfx) }$ where $p$ is a certain polynomial family. 

Let us now illustrate this relationship by considering a natural generic strategy, namely performing Maximum-Likelihood Estimation via gradient-ascent. For that purpose, let $\bfx \sim \gamma_d$, and $\bfy = g( \theta_*^\T \bfx) + \xi $, where $\theta_* \in \mathcal{S}_{d-1}$ is the planted direction and $\xi$ is a Gaussian noise independent of $\bfx$. Given iid samples $\{ (\bfx_i, \bfy_i) \}_i$ from this model, the MLE estimator in the parametric class $ \theta \mapsto \bfy | \bfx = \mathcal{N}( g(\theta^\T \bfx), 1)$ is proportional to 
\begin{align}
    \hat{L}(\theta) &= \frac1n \sum_i (\bfy_i - g(\theta^\T \bfx_i) )^2~.
\end{align}
It is instructive to consider the population limit of this empirical landscape, given by 
\begin{align}
\label{eq:empirical_loss}
    L(\theta) &:= \E\sbr{ | g(\theta_*^\T \bfx) - g(\theta^\T \bfx) |^2} + \E\sbr{\xi^2}~.
\end{align}
Denoting $g_\theta(x) = g(\theta^\T x)$ and using the rotation symmetry of the Gaussian measure, this landscape is, up to a constant, equivalent to the correlation 
$\langle g_\theta, g_{\theta_*} \rangle_{\gamma_d}$. Clearly, this Gaussian inner product is only a function of two (correlated) scalar Gaussian variables, in the span of $\theta, \theta_*$. By parametrizing them as $z \sim \gamma_1$ and $\tilde{z} = m z + \sqrt{1-m^2} w$, with $z, w \sim \gamma_1$ independent and $m= \theta^\T  \theta_*$, we obtain that 
\begin{align}
\label{eq:correlbasic}
    \langle g_\theta, g_{\theta_*} \rangle_{\gamma_d} & = \langle g, \mathsf{A}_m g \rangle_{\gamma_1}~,  
\end{align}
where, for $m\in [-1, 1]$, $\mathsf{A}_m$ is the Ornstein-Uhlenbeck (OU) semigroup given by 
\begin{align}
        \mathsf{A}_m g(z) &= \E_w \sbr{ g(m z + \sqrt{1-m^2} w) }~.
\end{align}
This commutative semigroup jointly diagonalises in $L^2(\R, \gamma)$, with eigenfunctions given by \emph{Hermite} polynomials $\{ h_k \}_k$ satisfying 
$\mathsf{A}_m h_k = m^k h_k$. 
This orthogonal structure turns out to be sufficient to provide an explicit description of the geometry of the MLE landscape above. Indeed, by decomposing the link function $g \in L^2(\R, \gamma)$ using this Hermite basis, $g = \sum_k \alpha_k h_k$, we obtain 
\begin{align}
\label{eq:correl}
\langle g_\theta, g_{\theta_*} \rangle &= \sum_k \alpha_k^2 m^k~. 
\end{align}
This representation of the loss reveals two important properties: on the one hand, thanks to the rotational invariance of the gaussian measure, this high-dimensional landscape is in fact only a function of a single scalar, the \emph{overlap} $m = \theta^\T \theta_*$. Additionally, not only is this landscape secretly one-dimensional, it also has a particularly simple topology: 
thanks to the fact that $\ell(m) := \sum_k \alpha_k^2 m^k$ satisfies $\ell'(m) > 0$ for $m>0$,  
there is a single saddle point at the equator $\{ \theta; \, m = 0\}$ within the upper hemisphere $\{ \theta; m \geq 0\}$; in particular, the loss has a single maximiser (which is global) . From (\ref{eq:correl}), one can also deduce that the order of the equatorial saddle is given by the index $l^\star$ of the first non-zero coefficient $\alpha_k$ --- precisely the number of vanishing moments of $g$, and referred as the \emph{information exponent} of $g$ in \cite{arous2021online} (or \emph{order of degeneracy} in \cite{dudeja2018learning}).

As a result, one should expect that from a random initialisation $\theta_0 \sim \mathrm{Unif}(S_{d-1})$ satisfying $m_0 \geq 0$ (which holds with probability $1/2$), a local method such as (projected) gradient ascent should converge towards the `North pole' $\theta_*$ in the population limit. 
Beyond this qualitative behavior, the effect of the high-dimension is felt once one tries to quantify this convergence in presence of only a finite number of samples. Indeed, a typical initialisation $\theta_0$ has correlation $m_0 \simeq 1/\sqrt{d}$, indicating that the gradient dynamics will be initialized at a neighborhood of the saddle point --- the so-called \emph{mediocrity zone}. Now, we can view the empirical landscape (\ref{eq:empirical_loss}) as a `noisy' perturbation of $L$, in which 
$\| \nabla \hat{L}(\theta) - \nabla L(\theta) \| \simeq \sqrt{\frac{d}{n}}$ uniformly over $\theta$ under mild smoothness assumptions \cite{mei2017landscapeempiricalrisknonconvex}. The empirical landscape will thus be successfully optimized whenever the signal gradient `strength', of order $\simeq m_0^{l^\star -1}\simeq d^{-(l^\star -1)/2}$, dominates the empirical fluctuations, of order $\sqrt{d/n}$. In other words, whenever $n \gg d^{l^\star}$, gradient methods will successfully optimise the MLE objective. 

This analysis was put forward in the seminal works of Ben Arous, Gheissari and Jagannath \cite{arous2021online}, as well as Dudeja and Hsu \cite{dudeja2018learning}, using slightly better algorithms. Specifically, \cite{arous2021online} considered online SGD, which replaces the uniform gradient concentration with a sharper martingale analysis along the trajectory, leading to a slight improvement in the rate, to $O(d^{l^\star-1})$ \footnote{The cases $l^\star=1$ and $l^\star = 2$ yield rates of $O(d)$ and $O(d \log d)$ respectively. This rate follows by studying the growth of a polynomial ODE of the form $\dot{m} = (1-m^2) m^{l^\star -1}$, which behaves differently for $l^\star>2$.}. On the other hand, \cite{dudeja2018learning} show that only two steps of a "gradient iteration" suffices:
\begin{equation*}
  \hat{\bf \theta}_0 \stackrel{\bfG}{\longrightarrow} {\bf \hat\theta}_1 \stackrel{\bfG}{\longrightarrow} {\bf \hat\theta}_2~,
\end{equation*}
where
\begin{equation*}
  \theta \mapsto \bfG(\theta) := \frac{1}{\norm{\nabla\bfF_{l^\star}(\theta)}}_2 \nabla\bfF_{l^\star}(\theta)~,
\end{equation*}
and $\bfF_l$ is the orthogonal projection of the correlation onto the $l$-th harmonic, ie $\bfF_l(\theta) := \frac1n \sum_{i=1}^n \bfy_i h_l(\theta^\T\bfx_i) .$
We remark that the previous estimation procedure can be extended even in the setting where the information exponent $l^\star$ is unknown, via a correlation-based goodness-of-fit criterion \cite{dudeja2018learning}. 

Finally, these improvements culminated in Damian et al.~\cite{damian2023smoothing}, who used a landscape smoothing procedure first introduced in the physics literature \cite{biroli2020iron}, akin to the partial trace estimation from tensor methods \cite{hopkins2018statistical}. While the PHD estimator (\ref{eq:est_phd}) consists in extracting the principal eigenvectors of the matrix ${\bf \widehat{M}}$, Damian et al consider instead the empirical tensor 
\begin{align}
\label{eq:hermtens}
    {\bf \widehat{T}} := \frac1n \sum_i \bfy_i H_{l^\star}(\bfx_i)~,
\end{align}
where $H_k(x)$ is the $k$-th order Hermite tensor. 
In the single-index setting, one easily verifies that $\E\sbr{{\bf \widehat{T}}} \propto \theta_*^{\otimes l^\star}$ is a rank-one tensor, and thus one can view the estimation of $\theta$ as a non-iid version of Tensor PCA, where from the observed data one builds a tensor ${\bf \widehat{T}}$ with planted rank-one structure --- but where entries are correlated, as opposed to Tensor PCA. 
Nonetheless, \cite{damian2023smoothing} leverage the partial-trace estimator, an efficient spectral method that boosts the signal-to-noise ratio by averaging the tensor in directions where the signal is constant, to estimate the planted direction with a sample complexity of $O( d^{l^\star / 2})$, which turns out to be optimal amongst the class of correlation-based statistical query algorithms \cite{damian2022neural, abbe2023sgd}.

Finally, let us mention that the geometric picture brought by the information exponent is robust to small perturbations of the Gaussian data distribution. Indeed, \cite{bruna2023single} demonstrates that spherical symmetry is sufficient, and that distributions whose sliced Wasserstein distance is of order $O(1/\sqrt{d})$ from the Gaussian reference also result in an efficient MLE gradient estimation whenever the information exponent is at most $2$. 

\subsection{Leap Exponents for Multi-Index Models}
\label{sec:multiindexgauss}

The previous section outlined a framework for single-index models that exploits the Hilbertian structure of the correlation loss to obtain an explicit `decoupling' of the two ingredients of single-index models, namely the hidden direction $\theta_*$ and the link function $g$. In essence, this is achieved thanks to the joint diagonalization of the OU semigroup via Hermite polynomials. 

A natural question is then how to extend this framework to the general multi-index setting. Denoting again $g_W(x) = g(W^\T x)$ for $W \in \mathrm{Stiefel}(d, r)$,
the first step is to generalise the representation (\ref{eq:correlbasic}); it now writes 
\begin{align}
    \label{eq:correl_multi}
        \langle g_W, g_{W_*} \rangle_{\gamma_d} & = \langle g, \mathsf{A}_M g \rangle_{\gamma_r}~,  
\end{align}
where now $\mathsf{A}_M$ is the matrix semigroup given by 
\begin{align}
        \mathsf{A}_M g(z) &= \E_{w \sim \gamma_r} \sbr{ g(M z + \sqrt{I-M^\T M} w) }~,
\end{align}
defined for correlation matrices $M = W^\T W_* \in \mathbb{R}^{r \times r}$ such that $\| M \| \leq 1$. 
An important point to realize is that now $\{ \mathsf{A}_M \}_M$ is no longer commutative, and thus one cannot hope to recreate the satisfying Hermite orthonormal decomposition from the single-index setting straight away. 
Instead, the correlation $\langle g_W, \tilde{g}_{\tilde{W}} \rangle_{\gamma_d}$ between two multi-index functions is now expressed as 
\begin{align}
    \langle g_W, \tilde{g}_{\tilde{W}} \rangle_{\gamma_d} &= \sum_{\beta} \langle g, H_\beta(U) \rangle \langle \tilde{g}, H_\beta(V) \rangle \prod_{j=1}^r \lambda_j^{\beta_j}~,
\end{align}
where $M = W^\T \tilde{W}= U \Lambda V^\T$ is the Singular-Value Decomposition (SVD) of $M$, $\beta=(\beta_1, \ldots, \beta_r) \in \mathbb{N}^r$ are multi-indices,  and $H_\beta(U)$ is the tensorised Hermite polynomial 
$H_\beta(U) (x) = \prod_{j=1}^r h_{\beta_j}( U_j^\T x)$ associated with the orthonormal basis $U$ of $\R^r$. 

While the squared-loss still admits a representation in terms of low-dimensional summary statistics $M \in \R^{r \times r}$, the monotonicity of the gradients is now lost: the effect of subspace internal misalignments (captured by the fact that $U \neq V$) can create spurious maxima in the MLE landscape. 
Such difficulty can be overcome by considering instead a two-timescale bilevel algorithm \cite{bietti2023learning}, that learns the link function $g$ in a non-parametric class of $r$-dimensional functions in the inner loop. This effectively replaces the correlation matrix $M$ by its Grammian $G = M^\T M$, leading to a `symmetrised' optimisation landscape  
\begin{align}
    L(W) &= 2 \|g\|^2 - 2 \sum_{\beta} \langle g, H_\beta(V) \rangle^2 \prod_{j=1}^r \lambda_j^{2\beta_j}~
\end{align}
that recovers the benign geometrical properties of its single-index counterpart. 

In particular, \cite{abbe2023sgd, bietti2023learning} show that local gradient methods on this landscape evolve along \emph{saddle-to-saddle} dynamics, as opposed to the single `escape from mediocrity saddle' of the single-index setting. The dynamics are still initialized at a neighborhood of a saddle point,  satisfying $\|M_0\|\simeq 1/\sqrt{d}$. Next, instead of directly escaping towards the global minimiser $W=W_*$, the gradient dynamics will visit a (finite) sequence of saddle points before reaching the global optimum. Remarkably, one can characterise these intermediate saddle points in terms of a hierarchical decomposition of the link function based on multivariate Hermite expansions. Informally speaking, the span of the lowest `frequencies' of $g$ is learnt first; subsequently, the gradient flow dynamics reveal the directions in the orthogonal complement carrying the lowest remaining frequencies, and so on. Each step in this sequential decomposition is characterized by a \emph{leap dimension} and \emph{leap exponent}, corresponding respectively to the index and the order of the associated saddle point. 

An important special setting where this sequential decomposition is particularly accessible is given by \emph{staircase} functions, first introduced by Abbe et al.~in \cite{abbe2021staircasepropertyhierarchicalstructure} in the context of Boolean functions (see also the earlier work of \cite{kalai2009learning} in a related context). In short, a function has the staircase property if the support of their high-frequency coefficients can be `reached' from the support of low-frequencies by adding a single coordinate, e.g., $g(z) = z_1 + z_1 z_2 + z_1 z_2 z_4$, for $z \in \{\pm 1\}^d$. 
The staircase property was further developed in \cite{abbe2022merged, abbe2023sgd}, where a sample complexity of order $d^{l^\star-1}$ is established for a certain class of generalised staircase functions, where $l^\star$ is the largest leap exponent (by slightly abusing notation), using local gradient methods on a suitable NN architecture. \cite{bietti2023learning} shows that $l^\star$ indeed controls the complexity in the general setting, focusing on population dynamics. 

Finally, another important instance of multi-index models with explicit algorithmic performance is given by \emph{committee machines}, corresponding to a link function of the form $g(z) = \text{sign}\left( \sum_{j=1}^r \text{sign}(z_j) \right)$ \cite{Aubin_2019}. In this setting, and in the language of the leap exponents, one can verify that there is energy in the linear harmonic, and thus there is no saddle-point structure, leading to a guarantee of the form $n = \Theta(d)$. Such guarantees were recently extended to a broader class of multi-index functions having non-zero linear terms in \cite{dandi2023twolayer}. 

In conclusion, algorithms based on local gradient ascent on the correlation are shown to be a powerful tool to estimate the index space, with sample complexity driven by the number of vanishing Hermite moments. One can now wonder to what extent these correlation methods are `optimal', at least in terms of said sample complexity.

\subsection{Beyond Information and Leap Exponents via Inverse Regression}
\label{sec:beyond_info}
As it turns out, correlation-based algorithms are not optimal in the previous sense. 
One of the clearest instances of this lack of optimality is the seminal work of Chen and Meka \cite{chen2020learning}: they showed that when $g$ is a $r$-dimensional, degree-$s$ polynomial, there exists an efficient algorithm that only needs
\begin{equation*}
  n = O_{r,s}( d \log^2(1/\epsilon) (\log d)^s)
\end{equation*}
samples to recover the subspace up to error $\epsilon$ in a noise-free setting.

The sample complexity is thus \emph{linear} in the dimension, matching the optimal information-theoretic rate. Taking for instance $g=h_s$ above shows an inherent advantage of this method over the previous correlation-based methods, which require $\Theta(d^{s/2})$ samples.  In essence, Chen and Meka exploit the \emph{inverse regression} idea, first put forward in Li's seminal work \cite{li1991sliced}: rather than focusing solely in the regression task of modeling $\bfy | \bfx$, one can gain information by looking instead at the inverse (or `generative') model $\bfx | \bfy$.  
The key observation is that, under Gaussian assumptions on $\bfx$, the moments $\E[\bfx^{\otimes k}| \bfy=y]$, seen as tensor-valued functions of $y$, are only non-constant within the index space.\footnote{Similar results hold under slightly milder conditions on $\bfx$; see \cite{klock2020estimating}.} This idea has led to several algorithmic solutions that exploit this moment structure \cite{Cook2000SaveAM, li2007directional, klock2020estimating}.

Another seemingly separate line of work, notably \cite{mondelli2018fundamental, barbier2019optimal,maillard2020phase}, aimed to characterize the class of single-index models such that one can efficiently recover the hidden direction $\theta$ in the so-called \emph{proportional regime}, where the number of samples satisfies $n = \alpha d$ with fixed $\alpha>0$, and $d$ goes to infinity. 
Specifically, these authors showed that whenever the joint distribution $(\bfx, \bfy)$ satisfies 
\begin{align}
\label{eq:gen2cond}
    \max\left\{ \E\sbr{ \E\sbr{ (\theta_*^\T \bfx)^2 - 1 | \bfy }^2 }, \E\sbr{ \E\sbr{ \theta_*^\T \bfx | \bfy }^2 }\right\} &> 0~,
\end{align}
then there are efficient algorithms that estimate $\theta$ provided $n \gtrsim d$. We identify the flavor of inverse regression in \eqref{eq:gen2cond}, through the conditioning on the label $\bfy$. 
Let us add that these works go beyond this guarantee, and even identify fundamental thresholds $\alpha_{\mathrm{I}} \leq \alpha_{\mathrm{C}}$ capturing the required sample complexity for brute-force and efficient estimators, respectively \footnote{These results are an instance of the so-called \emph{Computational-Statistical Gaps}, which study the interplay between statistical and computational aspects in high-dimensional inference. In particular, some inference tasks are known to exhibit a phase diagram with three distinct regimes: for $n \ll n_{\mathrm{I}}$, no estimator can recover the signal of interest, for $n \gg n_{\mathrm{C}}$, one can exhibit efficient estimation procedures that succeed, but for $n_{\mathrm{I}} \ll n \ll n_{\mathrm{C}}$, it is conjectured that \emph{no} efficient procedure can exist. Such conjecture is often formalized by restricting the class of estimation procedures to belong to a certain computational class, e.g., Statistical Queries \cite{kearns1998efficient}, or Low-degree polynomials \cite{hopkins2016fast} }.
In particular, \cite{lu2020phase,mondelli2018fundamental} considered a variant of the PHD estimator, where $\hat{{\bf \theta}}$ is the principal eigenvector of 
\begin{align}
\label{eq:mondelli}
    \widetilde{{\bf M}} &= \frac1n \sum_i \mathcal{T}(\bfy_i) ( \bfx_i \bfx_i^\T - I_d)~,
\end{align}
and where $\mathcal{T}:\R \to \R$ is a suitable label transformation. On the other hand, \cite{barbier2019optimal} developed an approach based on Approximate-Message-Passing (AMP) under the same assumption (\ref{eq:gen2cond}), in which there is also an underlying label transformation --- as is also the case in the aforementioned \cite{chen2020learning}. Finally, in \cite{maillard2022construction} it was shown that the spectral estimator associated with \eqref{eq:mondelli} is the linearisation of the AMP through the so-called Bethe Hessian.

In retrospect, it should not be surprising that a label transformation is able to overcome the information exponent barriers from the previous section; indeed, 
recalling that $l^\star(\mathsf{G}) = \min\{ l; \E\sbr{\bfy h_l(\bfx)} \neq 0\}$, one can easily verify that this definition is not invariant to compositions, ie it can happen that the change of variables $\tilde{\mathsf{G}} := (\mathrm{Id} \otimes \mathcal{T})_\# \mathsf{G}$, where $(\bfx, \bfy)$ is mapped to $(\bfx, \mathcal{T}(\bfy))$, satisfies $l^\star( \tilde{\mathsf{G}}) < l^\star(\mathsf{G})$! 

Such lack of composition invariance suggests that one should revisit the information exponent definition to ensure it cannot be decreased by label transformation, and thus that it captures the intrinsic difficulty of estimation beyond correlation-based methods. This can be achieved by considering 
\begin{align}
\label{eq:geninfo}
    k^\star(\mathsf{G}) &:= \inf_{\substack{\mathcal{T}:\R \to \R\\  \text{measureable} }} l^\star( \tilde{\mathsf{G}} ) ~.
\end{align}
This defines a new \emph{generative exponent } of the single-index model \cite{damian2024computational}. By simple consequence of (\ref{eq:geninfo}), we have that $k^\star( \mathsf{G}) \leq l^\star(\mathsf{G})$. 

This exponent can also be viewed as the leading term of an expansion. While the information exponent appears as the leading order term of the expansion of the $L^2$ energy $\E\sbr{\E\sbr{\bfy| \bfx}^2} = \| g \|^2 = \sum_l \E\sbr{\bfy h_l(\bfx)}^2 
=  \sum_{l \geq l^\star} \E\sbr{\bfy h_l(\bfx)}^2~$ (and thus associated with correlation), the generative exponent is associated with an expansion of another metric, given by the $\chi^2$ information: assuming that $\mathsf{G} \ll \mathsf{G}_0$, with $\mathsf{G}_0=\mathsf{G}_z \otimes \mathsf{G}_y$, ie that $\mathsf{G}$ is absolutely continuous with respect to the product of its marginals, the $\chi^2$ information is defined as $I( \mathsf{G}) = D_{\chi^2}( \mathsf{G} || \mathsf{G}_0) = \left\| \frac{\mathrm{d}\mathsf{G} }{ \mathrm{d}\mathsf{G}_0}\right \|^2_{\mathsf{G}_0} -1 $. 
Defining for each $k$ the conditional moment $\zeta_k(y) = \E\sbr{h_k(\bfz) | \bfy=y}$, one can verify \cite{damian2024computational} that 
\begin{align}
\label{eq:chi_info}
    I( \mathsf{G}) &= \sum_{k} \| \zeta_k \|^2_{\mathsf{G}_y} 
                    = \sum_{k \geq k^\star} \| \zeta_k \|^2_{\mathsf{G}_y}~.
\end{align}
We now recognize that (\ref{eq:gen2cond}) corresponds to the assumption $k^\star \leq 2$. 
The conditional moments in $\zeta_k$ fully realize the vision of Li's Inverse Regression: indeed, we now extract \emph{all} the moments of the conditional distribution $\bfx | \bfy$ rather than focusing only on the first few moments. 

The representation (\ref{eq:chi_info}) leads to a natural generalisation of the $L_2$ decomposition (\ref{eq:correl}) in terms of the likelihood ratios $\frac{\mathrm{d}\mathsf{\pi}_\theta }{ \mathrm{d}\mathsf{\pi}_0}$.
Equation~\eqref{eq:chi_info} thus reveals that the likelihood ratio is in the span of polynomials of degree at least $k^\star$ in $\bfx$. As a consequence, one can deduce \cite{damian2024computational} a lower bound for the required sample complexity of $n \gtrsim d^{k^\star/2}$, both under the SQ (Statistical-Query) and the Low-degree polynomial frameworks. These computational classes encompass a wide range of algorithms, including spectral and general moment methods, and (stochastic) gradient descent. 

Moreover, the function $\zeta_k(y)$ defines the optimal label transformation $\mathcal{T}$ that reduces the generative to the information exponent. Therefore, the natural confluence of (\ref{eq:hermtens}) and (\ref{eq:mondelli}) is to consider the following tensor:
\begin{align}
    {\bf \widetilde{T}} := \frac1n \sum_i \zeta_{k^\star}(\bfy_i) H_{k^\star}(\bfx_i)~.
\end{align}
Indeed, we verify that, again, this tensor has a planted spike precisely given by $\theta_*$, ie $\E\sbr{{\bf \widetilde{T}}} \propto \theta_*^{\otimes k^\star}$. 
As a result, one can establish \cite{damian2024computational} that a partial trace estimator will successfully estimate $\theta_*$ whenever $n \gtrsim d^{k^\star/2}$, thus tightly matching the computational lower bounds. As with previous algorithms, one can also extend this method to misspecified link distributions $\mathsf{G}$.

Equation~\eqref{eq:gen2cond} has been recently extended to the multi-index setting in \cite{troiani2024fundamentallimitsweaklearnability}, where AMP-based methods are shown to (weakly)\footnote{Weak recovery refers to an estimate $\hat{W}$ such that $\lambda_{\min}( \hat{W}^\top W_*) = \Theta(1)$, recalling that a subspace $W_0$ drawn uniformly at random satisfies $\lambda_{\min}( \hat{W}^\top W_*) \simeq 1/\sqrt{d}$.} recover the range of the following matrix in the proportional regime $n = \Theta(d)$:
\begin{align}
    & W_* \E\sbr{\zeta_1(\bfy) \zeta_1(\bfy)^\T + \zeta_2(\bfy)\zeta_2(\bfy)^\T} ~, \text{ with} \\
    & \zeta_1(y) = \E\sbr{H_1(\bfz) | \bfy=y}~,~\zeta_2(y) = \E\sbr{H_2(\bfz) | \bfy=y}~. \nonumber
\end{align}
Similarly, \cite{kovavcevic2025spectral} recently extended the analysis of \cite{mondelli2018fundamental} to the multi-index setting, establishing weak recovery of at least one direction inside the above index space using a spectral method.  

Finally, let us mention that lower bounds for multi-index models against several computational classes, including SQ and differentiable queries have been recently established in \cite{joshi2024complexitylearningsparsefunctions} for product measures admitting a predefined basis aligned with the unknown subspace, and in \cite{damian2025generative} for Gaussian data for the low-degree polynomial framework. 
Notably, these SQ/Low-degree lower bounds identify an appropriate extension of the generative exponent to the multi-index setting --- the \emph{leap generative} exponent. By adapting the partial trace estimator to this setting, \cite{damian2025generative} provide an algorithm matching again this sample complexity, therefore providing a `complete' statistical/computational description of the multi-index estimation problem in the high-dimensional regime (under Gaussian data); see also~\cite{diakonikolas2025algorithmssqlowerbounds,diakonikolas2025robustlearningmultiindexmodels} for concurrent works that establish both an SQ lower bound, and also provide an upper bound using a subspace conditioning algorithm.

\section{Gradient span}
\label{sec:gradients}

One of the critical drawbacks of the moment methods in \Cref{sec:moments} is the normality assumption on $\bfx$.
A standard approach to addressing this limitation is to employ score functions for non-normal distributions on $\bfx$~\citep[e.g.,][]{yang2017high}, but otherwise the techniques are mostly unchanged.
Some moment methods have also been shown to correctly estimate a subspace of the index space under weaker assumptions~\citep[e.g.,][]{li1989regression}, but these conditions are still fairly restrictive, and exhaustiveness is also only established under rather \emph{ad hoc} assumptions.

In this section, we describe a rather different approach based on using the span of gradients, which does not rely on the normality assumption and ensures exhaustiveness under fairly weak assumptions.
The primary drawback of this method is the difficulty of efficient estimation in high dimensions.

\subsection{Exhaustiveness of the gradient span}

The key idea of using the gradient span comes from the chain rule (as already exploited in \Cref{sec:moments}): the gradient of the regression function $\nabla\reg$ can be written as
\begin{equation*}
  \nabla\reg(x) = \lmap^\T \nabla\link(\lmap x) ,
\end{equation*}
so for any $x \in \R^d$, we have $\nabla\reg(x) \in W := \ran(\lmap^\T)$.
This suggests a natural strategy for estimating (at least a subspace of) $W$: estimate the gradient $\nabla\reg$ at several points, and then fit an $r$-dimensional subspace to these vectors.
Or more simply: estimate the
\emph{expected gradient outer product (EGOP)}\footnote{Also called \emph{Outer Product of Gradients (OPG)}~\citep{xia2002adaptive}.}
\begin{equation} \label{eq:egop}
  \EGOP := \E\sbr{\nabla\reg(\bfx)\nabla\reg(\bfx)^\T} = \lmap^\T \E\sbr{\nabla\link(\lmap\bfx) \nabla\link(\lmap\bfx)^\T} \lmap ,
\end{equation}
and then take the subspace spanned by the eigenvectors corresponding to the top $r$ eigenvalues of the estimate.

The gradient span approach guarantees exhaustiveness under fairly weak assumptions~\citep{samarov1993exploring}.
If there is a nonzero vector $w \in W$ orthogonal to $\nabla\reg(\bfx)$ almost surely, then $(\lmap w)^\T \nabla\link(\lmap \bfx) = 0$ almost surely.
This means that the link function $\link$ does not vary in the direction of $v := \lmap w$, contradicting the minimality of $W$ (recalling the definition of CMS)\footnote{%
  To make this argument precise, it suffices to assume convexity of the support of $\lmap\bfx$.%
}.
Therefore, we must have $W = \ran\del{\EGOP}$.
The exhaustiveness of (the range of) the EGOP makes it a natural target of estimation: a consistent estimate of $\EGOP$ yields a consistent estimate of $W$.

\subsection{Plug-in estimation}

A plug-in approach to estimating the EGOP is:
\begin{enumerate}
  \item Estimate the evaluation of $\nabla\reg$ on an i.i.d.~sample $\bfx_1,\dotsc,\bfx_n$ (perhaps using separate data independent of these $n$ points).
  \item Form the empirical average the outer products of the estimates of $\nabla\reg(\bfx_i)$:
    \begin{equation*}
      \frac1n \sum_{i=1}^n \widehat{\nabla\reg}(\bfx_i) \widehat{\nabla\reg}(\bfx_i)^\T .
    \end{equation*}
\end{enumerate}

To continue the plug-in stratregy, one can estimate $\nabla\reg$ by constructing an estimate $\hat\reg$ of $\reg$, and then using $\nabla\hat\reg$ as an estimate for $\nabla\reg$.
For example, \cite{samarov1993exploring} instantiates this approach with Nadaraya-Watson kernel regression methods~\citep{nadaraya1964estimating,watson1964smooth}, and \cite{kang2022forward} (as with \cite{mukherjee2006learning} in passing) do so with Reproducing Kernel Hilbert Space (RKHS) methods.
In both cases, the estimator $\hat\reg$ has an analytic form for which the gradient can be explicitly computed.

\subsection{Local estimates of gradients}
\label{sec:local_estimates_of_gradients}

\cite{hristache2001structure} suggest local linear regression~\citep{stone1977consistent} for estimating $\nabla\reg(x)$ in the present context.
Specifically, the estimate of the gradient at $x \in \R^d$ is
\begin{equation*}
  \widehat{\nabla\reg}(x) := b(x)
\end{equation*}
where $b(x) \in \R^d$ is obtained from a weighted linear least squares fit to an i.i.d.~sample $(\tilde\bfx_1,\tilde\bfy_1),\dotsc,(\tilde\bfx_{\tilde{n}},\tilde\bfy_{\tilde{n}})$:
\begin{equation*}
  (a(x),b(x)) :=
  \argmin_{(a,b) \in \R \times \R^d}
  \sum_{i=1}^{\tilde{n}} \bfw_i(x) \del{a + b^\T(\tilde\bfx_i - x) - \tilde\bfy_i}^2
\end{equation*}
with weights $\bfw_i(x) = K(\tilde\bfx_i,x)$ for some kernel function $K \colon \R^d \times \R^d \to \R$ (e.g., a radial kernel like $K(x,x') = \exp(-\norm{x-x'}_2^2/h^2)$ for a suitable bandwidth $h>0$).
Related estimation strategies based on Nadaraya-Watson kernel methods, RKHS methods and finite differences were proposed and analyzed, respectively, by \cite{samarov1993exploring}, \cite{mukherjee2006learning}, and \cite{trivedi2014consistent}.
The main downside of these approaches is that they suffer from the usual curse of dimension in nonparametric regression~\citep{stone1980optimal}: exponentially many (in $d$) data may be needed for accurate estimation.
\cite{samarov1993exploring} notes, however, that estimating the EGOP may be easier than estimating the gradient field, and gives conditions under which consistency at the parametric rate may be achieved \citep[see also][]{yuan2023efficient}.

\cite{hristache2001structure} (building on earlier work in the single-index case by~\cite{hristache2001direct}) show how a pilot estimate of the EGOP can be used to avoid the curse of dimension in a refined estimate.
In particular, in the refined estimate, the (elliptical) smoothing kernel $K$ is chosen so as to over-smooth in directions that the pilot estimate suggests are orthogonal to $W$.
Remarkably, this is enough to overcome the curse of dimension at least in the rate of convergence for the refined estimate: an exponential dependence on the ambient dimension $d$ is replaced by an exponential dependence only on the intrinsic dimension $r = \dim(W)$.
In fact, the rate is shown to be the parametric rate when $r \leq 3$ under some mild assumptions on the link function.

\subsection{Practical implementation and other settings}

The recent work of
\cite{radhakrishnan2024mechanism} combines the RKHS approach of \cite{kang2022forward} and the refinement technique of \cite{hristache2001structure} to obtain a practical algorithm.
Specifically, they employ RKHS-based regression with an isotropic Gaussian or Laplace kernel to for an initial estimate $\hat\reg$ of $\reg$, and then use $\nabla\hat\reg$ to estimate the EGOP as above.
Then the estimate of the EGOP is used to change the isotropic kernel to an anisotropic kernel in a manner very similar to that suggested by \cite{hristache2001structure} to get updated estimates of $\reg$ and also of the EGOP.
They suggest repeating this process several times to obtain a final estimate of $\reg$ and/or the EGOP.

Manual inspection of the final estimated EGOP in several datasets shows the ability of this approach to identify interpretable "features" that are intuitively relevant for the prediction task at hand.
For example, on a dataset for predicting whether an image depicts a smiling face or non-smiling face, the top eigenvectors of the EGOP estimate emphasize pixel features that roughly correspond to those where the eyes and mouth are located in the image.
The EGOP estimate obtained this way is also compared to the first layer weight matrices of deep neural networks trained using gradient descent on the same data, and the similarities are striking enough for the authors to posit a hypothesis connecting these methods.

Another remarkable non-parametric approach is the recent work \cite{follain2024nonparametric}; the authors replace the gradient outer product estimation by a more general non-parametric method based on regularized empirical risk minimization, that promotes the `sparsity' of the estimated function gradients using a multivariate Hermite expansion, under minimal distributional assumptions. 

Finally, let us mention \cite{hemant2012active}, in which the authors replace the iid sampling assumption by an active learning strategy to estimate the index space, leading to a sample complexity of order $O(d^2)$.
In the active (or query) learning setting, it is possible to adaptively sample $\bfy$ from its conditional distribution given $\bfx = x$ for any $x$.
This naturally gives a path towards avoiding the curse of dimension suffered by local linear regression and other non-parametric methods (cf.~\cite{hristache2001direct,hristache2001structure} and Section~\ref{sec:local_estimates_of_gradients}).
Other active learning methods for learning or testing single-index and multi-index models include~\citep{de2019your,de2021robust,gajjar2024agnostic,diakonikolas2024agnostically}.

\section{Neural Networks}
\label{sec:NN}

We have so far discussed dedicated algorithms that exploit the specific
structure of multi-index models. 
An alternative line of work considers instead a `generic' learning algorithm, based on gradient-descent on a differentiable model, e.g., a neural network: given $n$ iid samples $\{ (\bfx_i,\bfy_i) \}_{i \leq n}$ of a multi-index model $P$, a differentiable function $f_\theta: \bfx \mapsto f_\theta(\bfx)$, and a point-wise loss $\ell(\cdot, \cdot)$, the Empirical-Risk-Minimisation (ERM) consists of 
\begin{equation}
\label{eq:mseNN}
\min_\theta \frac1n \sum_i \ell( f_\theta(\bfx_i), \bfy_i)~.    
\end{equation}
A canonical choice of loss is the mean-squared error $\ell(x,y) = (x-y)^2$, 
which is amenable to powerful analysis techniques in $L^2(\R^d, \pi_x)$.
The general question in this context is thus to what extent the minimisers $\hat{\theta}$ of (\ref{eq:mseNN}) carry information on $W$. 
When $f_\theta$ is a Neural Network (NN) implementing a composition of the form $f_\theta(\bfx) = \Phi_{\Theta_2}( \Theta_1^\top \bfx)$ for a given $\Theta_1 \in \R^{d \times d'}$ and generic $\Phi$, a natural proxy for $\hat{W}$ is then the span of the first layer $\Theta_1$. As discussed earlier, the motivation in this case is to provide a quantitative description of \emph{feature learning}, namely the ability of the NN to automatically discover the relevant compositional structure for the task --- which in the case of multi-index models is directly given by the low-rank projection onto $W$. 

At first glance, it does not seem obvious that a generic gradient-based analysis of \eqref{eq:mseNN} should lead to any quantitative guarantees, given the non-convexity of the associated energy landscape, even for the simplest shallow NNs. Faced with this difficulty, \cite{mei2018mean, chizat2018global, rotskoff2018parameters, sirignano2020mean} put forward a mean-field formulation of such shallow learning. By expressing the optimization problem in terms of the empirical measure over parameters, one obtains a `lifted' functional in the space of measures which is convex. Such convexity can then be leveraged to establish global convergence guarantees of gradient dynamics in the infinitely-wide limit. In the setting where the data is generated by a multi-index model, \cite{hajjar2023symmetriesdynamicswidetwolayer} showed that the associated dynamics become dimension-free, as with the summary statistics we saw in \Cref{sec:infoexpo}. However, while providing important qualitative insights about feature learning, an important limitation of these mean-field formulations is their lack of quantitative guarantees as one tries to operate with finite-width networks.

An alternative route is then to `face' the non-convexity of the optimization problem and attempt to exploit structural properties to overcome the worst-case scenarios. 
Indeed, in Section \ref{sec:infoexpo} we observed that the MSE loss $L(\theta) = \| g_\theta - g_{\theta^\star} \|^2$ can be efficiently optimized with gradient methods from a random initialization. One can view such model as a contrived neural network consisting of a single neuron, $x \mapsto g( \theta^\top x)$. 
One could thus hope that gradient-based learning on a more general NN might be able to estimate the index space. Let us now review the main results in this direction.

\subsection{Information Exponent and NNs}

One of the early attempts that demonstrated the ability of NNs to estimate the index space is \cite{chen2021understandinghierarchicallearningbenefits}. Focusing on the setting where $g$ is a certain class of degree-$s$ polynomial, the authors show that a partially-trained 3-layer neural network can learn the target function with sample complexity $n = \widetilde{O}( d^{\lceil s / 2 \rceil})$, consistent with the CSQ-optimal rate of \cite{damian2023smoothing}. 
More generally, under the Hessian non-degeneracy assumption from \Cref{sec:hessian}, and assuming Gaussian inputs, \cite{damian2022neural,ba2022high,dandi2023twolayer,cui2024asymptotics} consider a shallow NN $f_{\theta}(x) = a^\top \sigma( W x + b)$ and squared-loss, and show that a \emph{single} gradient step with respect to the input weights $W$ is aligned with $W^\star$. Indeed, picking $\sigma(u) = \max(0, u)$ and denoting $W=[w_1, \ldots, w_m]$, the population gradient of each weight vector $w_j$ writes 
$$\nabla_{w_j} \| f_{\theta} - g_{W^\star} \|^2 \propto \E\left[g_{W^\star}(\bfx)\bfx \mathbf{1}(\bfx \cdot w_j \geq 0) \right]~.$$
Expanding again this inner product in an orthonormal Hermite basis and identifying the leading order term yields \cite[Lemma 1]{damian2022neural}
$$\nabla_{w_j} \| f_{\theta} - g_{W^\star} \|^2 = -\sqrt{2/\pi} a_j \E\sbr{\nabla^2 g_{W^\star}} w_j + \widetilde{O}(\sqrt{r}/d))~.$$
In words, the gradient at initialization reveals the Principal Hessian Directions. 
Under the assumption that these directions span the whole index space --- which in the language of \Cref{sec:multiindexgauss} corresponds to information exponent $l^\star \leq 2$ and a single leap, this turns out to be sufficient to establish a learning guarantee of order $n = O(d^2)$, via a hybrid gradient-descent strategy whereby second-layer features are kept frozen at initialization. By making the stronger assumption that $l^\star =1$, one can get sharper rates of $n=O(d)$  \cite{ba2022high, mousavi2022neural}, in accordance with \Cref{sec:infoexpo}. 

Beyond the setting of $l^\star \leq 2$, several works also analyse the performance of several NN architectures with gradient-based learning. As mentioned in \Cref{sec:multiindexgauss}, Abbe, Boix-Adsera and Misiakiewicz study in \cite{abbe2022merged,abbe2023sgd} the ability of shallow NNs to learn a class of multi-index models characterized by the \emph{staircase} property. In essence,  multi-index functions $g$ with the staircase property (or the `merged' staircase, an extension put forward in \cite{abbe2022merged}) admit a spectral decomposition in a certain Hermite basis $H_\beta(V)$ whereby the support of each harmonic is obtained from the lower harmonics by adding additional coordinates. As in \cite{damian2022neural, dandi2023twolayer, bietti2022learning}, the analysis of gradient-descent follows a similar layer-wise strategy, with the second-layer weights frozen at initialization, resulting in a sample complexity guarantee of the form $n=\widetilde{O}( d^{L^\star})$, where $L^\star$ is the \emph{leap} complexity of the model, ie the largest number of revealed coordinates when going from one harmonic to the next. 

Besides learning the hidden direction, another important question is to understand how the link function is being learnt. In \cite{berthier2024learningtimescalestwolayersneural}, the authors study standard shallow NNs in the mean-field regime, and demonstrate that the network naturally learns different harmonics of the target link function at different timescales. 
Alternatively, \cite{bietti2022learning} focuses on single-index models, and considers a `specialized' shallow NN $f_\theta(x) = a^\top \sigma( W x + b)$, with $W$ constrained to be a rank-one matrix of the form $W= {\bf 1} w^\top$, and frozen biases drawn from a normal distribution. In other words, neurons share their input weights and only differ by a random (and frozen) bias. By adapting the analysis of \cite{arous2020online, dudeja2018learning}, the authors establish non-parametric rates for general link functions with a sample complexity $n = O(d^{l^\star})$ to learn both the index direction and the link function.
Remarkably, the efficient algorithm of \cite{chen2022deeprelu} uses an ingenious enumeration technique based on lattice polynomials to learn any link function specified by a (homogeneous) deep ReLU network.

More recently, \cite{dandi2024random} leveraged  random matrix theory to study the ability of a single gradient step on a shallow NN to identify the direction in a Single-Index model, in the high-dimensional proportional regime. 
Finally, while the majority of the analyses focus on Gaussian or Boolean inputs, 
some authors have explored the effect of anisotropy in the data \cite{mousavi2023gradient,ba2024learning}, highlighting the importance of symmetries in establishing learning guarantees. 

\subsection{Breaking the Information Exponent with NNs}
So far, all methods described require a sample complexity that scales with the information exponent. Based on the arguments from \Cref{sec:beyond_info}, it is natural to ask whether NNs can go beyond correlation-based learning and `break' the information-exponent barrier. 

As it turns out, the answer, put forward in \cite{Dandi2024TheBO, arnaboldi2024repetitaiuvantdatarepetition, lee2024neuralnetworklearnslowdimensional}, is yes. The key idea is that \emph{reusing} data batches to estimate empirical gradients is akin to performing a label transformation that, under appropriate scaling of the training hyperparameters, realises the lowering of the exponent from \eqref{eq:geninfo}. 
Indeed, by considering the correlation loss $\ell(\hat{y}, y) = 1 - \hat{y} y$ in \eqref{eq:mseNN}, and a shallow NN of the form $f_\theta(x) = a^\top \sigma( W x)$, 
the gradient with respect to each input weight $w_j$ evaluated at a single datapoint $(\bfx, \bfy)$ is 
\begin{align}
    \nabla_{w_j} \ell( f_\theta(\bfx), \bfy) &= -a_j \bfy \sigma'( w_j \cdot \bfx) \bfx ~.
\end{align}
Now, if we take a gradient step
\begin{equation*}
  \tilde{W} := W - \rho \nabla_{W} \ell( f_\theta(\bfx), \bfy)
\end{equation*}
and evaluate the gradient, using \emph{again} the same datapoint $(\bfx, \bfy)$, we obtain \cite[Eq 13]{arnaboldi2024repetitaiuvantdatarepetition}
\begin{multline*}
  \nabla_{w_j} \ell( f_{a, \tilde{W}}(\bfx, \bfy), \bfy) \\ = a_j \bfy \sigma'\left( w_j \cdot \bfx + a \rho \bfy \|\bfx \|^2 \sigma'(w_j \cdot \bfx)  \right) \bfx~.
\end{multline*}
An important property of this update is that now this gradient is outside the class of correlational queries, since it is a non-linear function of $y$. Since the generative exponent from \eqref{eq:chi_info} describes a `closed' property, namely that certain conditional moments $k < k^\star$ are zero, a generic choice of learning rate $\rho$ such that $\rho \| \bfx \|^2 \approx \rho d = \Theta(1)$ ensures that almost surely the resulting algorithm will now be driven by the generative exponent, rather than the information exponent of single-pass gradient methods. The resulting algorithm is analyzed in \cite{arnaboldi2024repetitaiuvantdatarepetition, lee2024neuralnetworklearnslowdimensional} in the setting where $k^\star \leq 2$ \footnote{Under a slightly more contrived defintion of generative exponent in \cite{arnaboldi2024repetitaiuvantdatarepetition}, where label transformations are limited to polynomials, and assuming polynomial link functions in \cite{lee2024neuralnetworklearnslowdimensional}}, leading to a linear sample complexity guarantee. 

In the context of multi-index models, the batch reuse is also shown to improve the sample complexities of the correlational queries framework. In particular, the staircase property characterizing efficiently learnable functions using correlation-based (S)GD via the leap exponent is now upgraded to the so-called `grand staircase' property \cite{troiani2024fundamentallimitsweaklearnability}. 

Another strategy to overcome the sample complexity barriers dictated by the information and generative exponents is to consider shallow neural networks in the mean-field regime \cite{mei2018mean,rotskoff2018parameters,chizat2018global,sirignano2020mean}, trained with noisy gradient descent, referred as mean-field Langevin dynamics. In this setting, \cite{mousavi2024learning} demonstrates information-theoretically optimal sample complexity to learn multi-index models, at the expense of exponential memory and/or runtime. 

Finally, let us mention that, thus far, all the works described above consider a certain layer-wise training scheme, whereby only first-layer weights move while the rest are frozen \cite{abbe2023sgd, bietti2022learning, arnaboldi2024repetitaiuvantdatarepetition, damian2022neural, lee2024neuralnetworklearnslowdimensional}, or by introducing timescale separation between layers \cite{bietti2023learning,berthier2024learningtimescalestwolayersneural}. 
A notable exception to this is the work of Glasgow \cite{glasgow2023sgdfindstunesfeatures}, which, focusing on Boolean data and the 2-parity $\mathrm{XOR}$ function, established that $n=\widetilde{O}(d)$ samples are sufficient to learn, using a standard shallow NN and standard mini-batch SGD training.

\bibliographystyle{alpha}
\bibliography{bibliography}       %

\newcommand{\etalchar}[1]{$^{#1}$}
\begin{thebibliography}{MHWSE23}

\bibitem[ABAB{\etalchar{+}}21]{abbe2021staircasepropertyhierarchicalstructure}
Emmanuel Abbe, Enric Boix-Adsera, Matthew Brennan, Guy Bresler, and Dheeraj Nagaraj.
\newblock The staircase property: How hierarchical structure can guide deep learning.
\newblock {\em arXiv preprint arXiv:2108.10573}, 2021.

\bibitem[ABAM22]{abbe2022merged}
Emmanuel Abbe, Enric Boix-Adsera, and Theodor Misiakiewicz.
\newblock The merged-staircase property: a necessary and nearly sufficient condition for sgd learning of sparse functions on two-layer neural networks.
\newblock {\em arXiv preprint arXiv:2202.08658}, 2022.

\bibitem[ABAM23]{abbe2023sgd}
Emmanuel Abbe, Enric Boix-Adsera, and Theodor Misiakiewicz.
\newblock Sgd learning on neural networks: leap complexity and saddle-to-saddle dynamics.
\newblock {\em arXiv preprint arXiv:2302.11055}, 2023.

\bibitem[ADK{\etalchar{+}}24]{arnaboldi2024repetitaiuvantdatarepetition}
Luca Arnaboldi, Yatin Dandi, Florent Krzakala, Luca Pesce, and Ludovic Stephan.
\newblock Repetita iuvant: Data repetition allows sgd to learn high-dimensional multi-index functions.
\newblock {\em arXiv preprint arXiv:2405.15459}, 2024.

\bibitem[AGJ21]{arous2021online}
Gerard~Ben Arous, Reza Gheissari, and Aukosh Jagannath.
\newblock Online stochastic gradient descent on non-convex losses from high-dimensional inference.
\newblock {\em Journal of Machine Learning Research}, 22(106):1--51, 2021.

\bibitem[AMB{\etalchar{+}}19]{Aubin_2019}
Benjamin Aubin, Antoine Maillard, Jean Barbier, Florent Krzakala, Nicolas Macris, and Lenka Zdeborová.
\newblock The committee machine: computational to statistical gaps in learning a two-layers neural network.
\newblock {\em Journal of Statistical Mechanics: Theory and Experiment}, 2019(12):124023, 2019.

\bibitem[BAGJ21]{arous2020online}
Gerard Ben~Arous, Reza Gheissari, and Aukosh Jagannath.
\newblock Online stochastic gradient descent on non-convex losses from high-dimensional inference.
\newblock {\em Journal of Machine Learning Research (JMLR)}, 22:106--1, 2021.

\bibitem[BBPV23]{bietti2023learning}
Alberto Bietti, Joan Bruna, and Loucas Pillaud-Vivien.
\newblock On learning gaussian multi-index models with gradient flow.
\newblock {\em arXiv preprint arXiv:2310.19793}, 2023.

\bibitem[BBSS22]{bietti2022learning}
Alberto Bietti, Joan Bruna, Clayton Sanford, and Min~Jae Song.
\newblock Learning single-index models with shallow neural networks.
\newblock {\em arXiv preprint arXiv:2210.15651}, 2022.

\bibitem[BC64]{https://doi.org/10.1111/j.2517-6161.1964.tb00553.x}
G.~E.~P. Box and D.~R. Cox.
\newblock An analysis of transformations.
\newblock {\em Journal of the Royal Statistical Society: Series B (Methodological)}, 26(2):211--243, 1964.

\bibitem[BCRT20]{biroli2020iron}
Giulio Biroli, Chiara Cammarota, and Federico Ricci-Tersenghi.
\newblock How to iron out rough landscapes and get optimal performances: averaged gradient descent and its application to tensor pca.
\newblock {\em Journal of Physics A: Mathematical and Theoretical}, 53(17):174003, 2020.

\bibitem[BD81]{66ace186-d33e-34c6-b587-c776d756007c}
Peter~J. Bickel and Kjell~A. Doksum.
\newblock An analysis of transformations revisited.
\newblock {\em Journal of the American Statistical Association}, 76(374):296--311, 1981.

\bibitem[BES{\etalchar{+}}22]{ba2022high}
Jimmy Ba, Murat~A Erdogdu, Taiji Suzuki, Zhichao Wang, Denny Wu, and Greg Yang.
\newblock High-dimensional asymptotics of feature learning: How one gradient step improves the representation.
\newblock {\em arXiv preprint arXiv:2205.01445}, 2022.

\bibitem[BES{\etalchar{+}}24]{ba2024learning}
Jimmy Ba, Murat~A Erdogdu, Taiji Suzuki, Zhichao Wang, and Denny Wu.
\newblock Learning in the presence of low-dimensional structure: a spiked random matrix perspective.
\newblock {\em Advances in Neural Information Processing Systems}, 36, 2024.

\bibitem[BKM{\etalchar{+}}19]{barbier2019optimal}
Jean Barbier, Florent Krzakala, Nicolas Macris, Léo Miolane, and Lenka Zdeborová.
\newblock Optimal errors and phase transitions in high-dimensional generalized linear models.
\newblock {\em Proceedings of the National Academy of Sciences}, 116(12):5451–5460, 2019.

\bibitem[BMZ24]{berthier2024learningtimescalestwolayersneural}
Raphaël Berthier, Andrea Montanari, and Kangjie Zhou.
\newblock Learning time-scales in two-layers neural networks.
\newblock {\em arXiv preprint arXiv:2303.00055}, 2024.

\bibitem[BPVZ23]{bruna2023single}
Joan Bruna, Loucas Pillaud-Vivien, and Aaron Zweig.
\newblock On single index models beyond gaussian data.
\newblock {\em arXiv preprint arXiv:2307.15804}, 2023.

\bibitem[Bri82]{brillinger1982generalized}
David~R. Brillinger.
\newblock A generalized linear model with `gaussian' regressor variables.
\newblock In P.~J. Bickel, K.~A. Doksum, and J.~A. Hodges, editors, {\em A Festschrift for Erich L. Lehmann}, pages 97--114. CRC Press, 1982.

\bibitem[CB18]{chizat2018global}
L{\'e}na{\"\i}c Chizat and Francis Bach.
\newblock On the global convergence of gradient descent for over-parameterized models using optimal transport.
\newblock In {\em Advances in Neural Information Processing Systems}, pages 3036--3046, 2018.

\bibitem[CBL{\etalchar{+}}21]{chen2021understandinghierarchicallearningbenefits}
Minshuo Chen, Yu~Bai, Jason~D. Lee, Tuo Zhao, Huan Wang, Caiming Xiong, and Richard Socher.
\newblock Towards understanding hierarchical learning: Benefits of neural representations.
\newblock {\em arXiv preprint arXiv:2006.13436}, 2021.

\bibitem[CKK{\etalchar{+}}24a]{chandrasekaran2024smoothed}
Gautam Chandrasekaran, Adam Klivans, Vasilis Kontonis, Raghu Meka, and Konstantinos Stavropoulos.
\newblock Smoothed analysis for learning concepts with low intrinsic dimension.
\newblock In {\em The Thirty Seventh Annual Conference on Learning Theory}, pages 876--922. PMLR, 2024.

\bibitem[CKK{\etalchar{+}}24b]{chandrasekaran2024efficient}
Gautam Chandrasekaran, Adam Klivans, Vasilis Kontonis, Konstantinos Stavropoulos, and Arsen Vasilyan.
\newblock Efficient discrepancy testing for learning with distribution shift.
\newblock {\em Advances in Neural Information Processing Systems}, 37:137263--137308, 2024.

\bibitem[CKM22]{chen2022deeprelu}
Sitan Chen, Adam~R. Klivans, and Raghu Meka.
\newblock Learning deep relu networks is fixed-parameter tractable.
\newblock In {\em 2021 IEEE 62nd Annual Symposium on Foundations of Computer Science (FOCS)}, pages 696--707, 2022.

\bibitem[CL02]{cook2002dimension}
R~Dennis Cook and Bing Li.
\newblock Dimension reduction for conditional mean in regression.
\newblock {\em The Annals of Statistics}, 30(2):455--474, 2002.

\bibitem[CM20]{chen2020learning}
Sitan Chen and Raghu Meka.
\newblock Learning polynomials in few relevant dimensions.
\newblock In {\em Conference on Learning Theory}, pages 1161--1227. PMLR, 2020.

\bibitem[Coo00]{Cook2000SaveAM}
R.~Dennis Cook.
\newblock Save: a method for dimension reduction and graphics in regression.
\newblock {\em Communications in Statistics-theory and Methods}, 29:2109--2121, 2000.

\bibitem[CPD{\etalchar{+}}24]{cui2024asymptotics}
Hugo Cui, Luca Pesce, Yatin Dandi, Florent Krzakala, Yue~M Lu, Lenka Zdeborov{\'a}, and Bruno Loureiro.
\newblock Asymptotics of feature learning in two-layer networks after one gradient-step.
\newblock {\em arXiv preprint arXiv:2402.04980}, 2024.

\bibitem[CR84]{7dc1b2c4-49df-3797-9492-fc8d6b538b08}
Raymond~J. Carroll and David Ruppert.
\newblock Power transformations when fitting theoretical models to data.
\newblock {\em Journal of the American Statistical Association}, 79(386):321--328, 1984.

\bibitem[DH18]{dudeja2018learning}
Rishabh Dudeja and Daniel Hsu.
\newblock Learning single-index models in gaussian space.
\newblock In {\em Thirty-First Annual Conference on Learning Theory}, 2018.

\bibitem[DH24]{dudeja2024statisticalcomputational}
Rishabh Dudeja and Daniel Hsu.
\newblock Statistical-computational trade-offs in tensor {PCA} and related problems via communication complexity.
\newblock {\em The Annals of Statistics}, 52(1):131--156, 2024.

\bibitem[DHK{\etalchar{+}}21]{du2021few}
Simon~Shaolei Du, Wei Hu, Sham~M Kakade, Jason~D Lee, and Qi~Lei.
\newblock Few-shot learning via learning the representation, provably.
\newblock In {\em International Conference on Learning Representations}, 2021.

\bibitem[DIKR25]{diakonikolas2025algorithmssqlowerbounds}
Ilias Diakonikolas, Giannis Iakovidis, Daniel~M. Kane, and Lisheng Ren.
\newblock Algorithms and sq lower bounds for robustly learning real-valued multi-index models.
\newblock {\em arXiv preprint arXiv:2505.21475}, 2025.

\bibitem[DIKZ25]{diakonikolas2025robustlearningmultiindexmodels}
Ilias Diakonikolas, Giannis Iakovidis, Daniel~M. Kane, and Nikos Zarifis.
\newblock Robust learning of multi-index models via iterative subspace approximation.
\newblock {\em arXiv preprint arXiv:2502.09525}, 2025.

\bibitem[DKK{\etalchar{+}}24]{diakonikolas2024agnostically}
Ilias Diakonikolas, Daniel~M Kane, Vasilis Kontonis, Christos Tzamos, and Nikos Zarifis.
\newblock Agnostically learning multi-index models with queries.
\newblock In {\em Sixty-fifth Annual Symposium on Foundations of Computer Science}, pages 1931--1952, 2024.

\bibitem[DKL{\etalchar{+}}23]{dandi2023twolayer}
Yatin Dandi, Florent Krzakala, Bruno Loureiro, Luca Pesce, and Ludovic Stephan.
\newblock How two-layer neural networks learn, one (giant) step at a time.
\newblock {\em arXiv preprint arXiv:2305.18270}, 2023.

\bibitem[DLB25]{damian2025generative}
Alex Damian, Jason~D Lee, and Joan Bruna.
\newblock The generative leap: Sharp sample complexity for efficiently learning gaussian multi-index models.
\newblock {\em arXiv preprint arXiv:2506.05500}, 2025.

\bibitem[DLS22]{damian2022neural}
Alexandru Damian, Jason Lee, and Mahdi Soltanolkotabi.
\newblock Neural networks can learn representations with gradient descent.
\newblock In {\em Conference on Learning Theory}, 2022.

\bibitem[DMN19]{de2019your}
Anindya De, Elchanan Mossel, and Joe Neeman.
\newblock Is your function low dimensional?
\newblock In {\em Conference on Learning Theory}, pages 979--993, 2019.

\bibitem[DMN21]{de2021robust}
Anindya De, Elchanan Mossel, and Joe Neeman.
\newblock Robust testing of low dimensional functions.
\newblock In {\em Fifty-third Annual Symposium on Theory of Computing}, pages 584--597, 2021.

\bibitem[DNGL23]{damian2023smoothing}
Alex Damian, Eshaan Nichani, Rong Ge, and Jason~D Lee.
\newblock Smoothing the landscape boosts the signal for sgd: Optimal sample complexity for learning single index models.
\newblock {\em arXiv preprint arXiv:2305.10633}, 2023.

\bibitem[DPC{\etalchar{+}}24]{dandi2024random}
Yatin Dandi, Luca Pesce, Hugo Cui, Florent Krzakala, Yue~M Lu, and Bruno Loureiro.
\newblock A random matrix theory perspective on the spectrum of learned features and asymptotic generalization capabilities.
\newblock {\em arXiv preprint arXiv:2410.18938}, 2024.

\bibitem[DPVLB24]{damian2024computational}
Alex Damian, Loucas Pillaud-Vivien, Jason~D Lee, and Joan Bruna.
\newblock Computational-statistical gaps in gaussian single-index models.
\newblock {\em arXiv preprint arXiv:2403.05529}, 2024.

\bibitem[DTA{\etalchar{+}}24]{Dandi2024TheBO}
Yatin Dandi, Emanuele Troiani, Luca Arnaboldi, Luca Pesce, Lenka Zdeborov'a, and Florent Krzakala.
\newblock The benefits of reusing batches for gradient descent in two-layer networks: Breaking the curse of information and leap exponents.
\newblock {\em arXiv preprint arXiv:2402.03220}, 2024.

\bibitem[FB24]{follain2024nonparametric}
Bertille Follain and Francis Bach.
\newblock Nonparametric linear feature learning in regression through regularisation.
\newblock {\em Electronic Journal of Statistics}, 18(2):4075--4118, 2024.

\bibitem[Gla23]{glasgow2023sgdfindstunesfeatures}
Margalit Glasgow.
\newblock Sgd finds then tunes features in two-layer neural networks with near-optimal sample complexity: A case study in the xor problem.
\newblock {\em arXiv preprint arXiv:2309.15111}, 2023.

\bibitem[GTX{\etalchar{+}}24]{gajjar2024agnostic}
Aarshvi Gajjar, Wai~Ming Tai, Xu~Xingyu, Chinmay Hegde, Christopher Musco, and Yi~Li.
\newblock Agnostic active learning of single index models with linear sample complexity.
\newblock In {\em Thirty-Seventh Annual Conference on Learning Theory}, pages 1715--1754, 2024.

\bibitem[HC12]{hemant2012active}
Tyagi Hemant and Volkan Cevher.
\newblock Active learning of multi-index function models.
\newblock {\em Advances in Neural Information Processing Systems}, 25, 2012.

\bibitem[HC23]{hajjar2023symmetriesdynamicswidetwolayer}
Karl Hajjar and Lenaic Chizat.
\newblock On the symmetries in the dynamics of wide two-layer neural networks.
\newblock {\em arXiv preprint arXiv:2211.08771}, 2023.

\bibitem[HJPS01]{hristache2001structure}
Marian Hristache, Anatoli Juditsky, Jorg Polzehl, and Vladimir Spokoiny.
\newblock Structure adaptive approach for dimension reduction.
\newblock {\em The Annals of Statistics}, 29(6):1537--1566, 2001.

\bibitem[HJS01]{hristache2001direct}
Marian Hristache, Anatoli Juditsky, and Vladimir Spokoiny.
\newblock Direct estimation of the index coefficient in a single-index model.
\newblock {\em The Annals of Statistics}, pages 595--623, 2001.

\bibitem[Hop18]{hopkins2018statistical}
Samuel Hopkins.
\newblock {\em Statistical inference and the sum of squares method}.
\newblock PhD thesis, Cornell University, 2018.

\bibitem[HS89]{hardle1989investigating}
Wolfgang H{\"a}rdle and Thomas~M Stoker.
\newblock Investigating smooth multiple regression by the method of average derivatives.
\newblock {\em Journal of the American Statistical Association}, 84(408):986--995, 1989.

\bibitem[HSSS16]{hopkins2016fast}
Samuel~B Hopkins, Tselil Schramm, Jonathan Shi, and David Steurer.
\newblock Fast spectral algorithms from sum-of-squares proofs: tensor decomposition and planted sparse vectors.
\newblock In {\em Proceedings of the forty-eighth annual ACM symposium on Theory of Computing}, pages 178--191, 2016.

\bibitem[JMS24]{joshi2024complexitylearningsparsefunctions}
Nirmit Joshi, Theodor Misiakiewicz, and Nathan Srebro.
\newblock On the complexity of learning sparse functions with statistical and gradient queries.
\newblock {\em arXiv preprint arXiv:2407.05622}, 2024.

\bibitem[Kea98]{kearns1998efficient}
Michael Kearns.
\newblock Efficient noise-tolerant learning from statistical queries.
\newblock {\em Journal of the ACM (JACM)}, 45(6):983--1006, 1998.

\bibitem[KLV20]{klock2020estimating}
Timo Klock, Alessandro Lanteri, and Stefano Vigogna.
\newblock Estimating multi-index models with response-conditional least squares.
\newblock {\em arXiv preprint arXiv:2003.04788}, 2020.

\bibitem[KOS08]{klivans2008learning}
Adam~R Klivans, Ryan O'Donnell, and Rocco~A Servedio.
\newblock Learning geometric concepts via gaussian surface area.
\newblock In {\em 49th Annual IEEE Symposium on Foundations of Computer Science}, pages 541--550, 2008.

\bibitem[KS22]{kang2022forward}
Jongkyeong Kang and Seung~Jun Shin.
\newblock A forward approach for sufficient dimension reduction in binary classification.
\newblock {\em Journal of Machine Learning Research}, 23(199):1--31, 2022.

\bibitem[KST09]{kalai2009learning}
Adam~Tauman Kalai, Alex Samorodnitsky, and Shang-Hua Teng.
\newblock Learning and smoothed analysis.
\newblock In {\em Fiftieth Annual Symposium on Foundations of Computer Science}, pages 395--404, 2009.

\bibitem[KSV24]{klivans2024learning}
Adam Klivans, Konstantinos Stavropoulos, and Arsen Vasilyan.
\newblock Learning intersections of halfspaces with distribution shift: Improved algorithms and sq lower bounds.
\newblock In {\em The Thirty Seventh Annual Conference on Learning Theory}, pages 2944--2978. PMLR, 2024.

\bibitem[KZM25]{kovavcevic2025spectral}
Filip Kova{\v{c}}evi{\'c}, Yihan Zhang, and Marco Mondelli.
\newblock Spectral estimators for multi-index models: Precise asymptotics and optimal weak recovery.
\newblock {\em arXiv preprint arXiv:2502.01583}, 2025.

\bibitem[LD89]{li1989regression}
Ker-Chau Li and Naihua Duan.
\newblock Regression analysis under link violation.
\newblock {\em The Annals of Statistics}, pages 1009--1052, 1989.

\bibitem[Li91]{li1991sliced}
Ker-Chau Li.
\newblock Sliced inverse regression for dimension reduction.
\newblock {\em Journal of the American Statistical Association}, 86(414):316--327, 1991.

\bibitem[Li92]{li1992principal}
Ker-Chau Li.
\newblock On principal hessian directions for data visualization and dimension reduction: Another application of stein's lemma.
\newblock {\em Journal of the American Statistical Association}, 87(420):1025--1039, 1992.

\bibitem[LL20]{lu2020phase}
Yue~M Lu and Gen Li.
\newblock Phase transitions of spectral initialization for high-dimensional non-convex estimation.
\newblock {\em Information and Inference: A Journal of the IMA}, 9(3):507--541, 2020.

\bibitem[LOSW24]{lee2024neuralnetworklearnslowdimensional}
Jason~D. Lee, Kazusato Oko, Taiji Suzuki, and Denny Wu.
\newblock Neural network learns low-dimensional polynomials with sgd near the information-theoretic limit.
\newblock {\em arXiv preprint arXiv:2406.01581}, 2024.

\bibitem[LW07]{li2007directional}
Bing Li and Shaoli Wang.
\newblock On directional regression for dimension reduction.
\newblock {\em Journal of the American Statistical Association}, 102(479):997--1008, 2007.

\bibitem[MBM17]{mei2017landscapeempiricalrisknonconvex}
Song Mei, Yu~Bai, and Andrea Montanari.
\newblock The landscape of empirical risk for non-convex losses.
\newblock {\em arXiv preprint arXiv:1607.06534}, 2017.

\bibitem[McC18]{mccullagh2018tensor}
Peter McCullagh.
\newblock {\em Tensor methods in statistics: Monographs on statistics and applied probability}.
\newblock Chapman and Hall/CRC, 2018.

\bibitem[MHPG{\etalchar{+}}22]{mousavi2022neural}
Alireza Mousavi-Hosseini, Sejun Park, Manuela Girotti, Ioannis Mitliagkas, and Murat~A Erdogdu.
\newblock Neural networks efficiently learn low-dimensional representations with sgd.
\newblock {\em arXiv preprint arXiv:2209.14863}, 2022.

\bibitem[MHWE24]{mousavi2024learning}
Alireza Mousavi-Hosseini, Denny Wu, and Murat~A Erdogdu.
\newblock Learning multi-index models with neural networks via mean-field langevin dynamics.
\newblock {\em arXiv preprint arXiv:2408.07254}, 2024.

\bibitem[MHWSE23]{mousavi2023gradient}
Alireza Mousavi-Hosseini, Denny Wu, Taiji Suzuki, and Murat~A Erdogdu.
\newblock Gradient-based feature learning under structured data.
\newblock {\em Advances in Neural Information Processing Systems}, 36:71449--71485, 2023.

\bibitem[MKLZ22]{maillard2022construction}
Antoine Maillard, Florent Krzakala, Yue~M Lu, and Lenka Zdeborov{\'a}.
\newblock Construction of optimal spectral methods in phase retrieval.
\newblock In {\em Mathematical and Scientific Machine Learning}, pages 693--720. PMLR, 2022.

\bibitem[MLKZ20]{maillard2020phase}
Antoine Maillard, Bruno Loureiro, Florent Krzakala, and Lenka Zdeborová.
\newblock Phase retrieval in high dimensions: Statistical and computational phase transitions.
\newblock {\em arXiv preprint arXiv:2006.05228}, 2020.

\bibitem[MM18]{mondelli2018fundamental}
Marco Mondelli and Andrea Montanari.
\newblock Fundamental limits of weak recovery with applications to phase retrieval.
\newblock In {\em Conference On Learning Theory}, pages 1445--1450. PMLR, 2018.

\bibitem[MMN18]{mei2018mean}
Song Mei, Andrea Montanari, and Phan-Minh Nguyen.
\newblock A mean field view of the landscape of two-layer neural networks.
\newblock {\em Proceedings of the National Academy of Sciences}, 115(33):E7665--E7671, 2018.

\bibitem[MZ06]{mukherjee2006learning}
Sayan Mukherjee and Ding-Xuan Zhou.
\newblock Learning coordinate covariances via gradients.
\newblock {\em Journal of Machine Learning Research}, 7(3), 2006.

\bibitem[Nad64]{nadaraya1964estimating}
Elizbar~A Nadaraya.
\newblock On estimating regression.
\newblock {\em Theory of Probability \& Its Applications}, 9(1):141--142, 1964.

\bibitem[PV12]{plan2012robust}
Yaniv Plan and Roman Vershynin.
\newblock Robust 1-bit compressed sensing and sparse logistic regression: A convex programming approach.
\newblock {\em IEEE Transactions on Information Theory}, 59(1):482--494, 2012.

\bibitem[RBPB24]{radhakrishnan2024mechanism}
Adityanarayanan Radhakrishnan, Daniel Beaglehole, Parthe Pandit, and Mikhail Belkin.
\newblock Mechanism for feature learning in neural networks and backpropagation-free machine learning models.
\newblock {\em Science}, 383(6690):1461--1467, 2024.

\bibitem[RVE18]{rotskoff2018parameters}
Grant Rotskoff and Eric Vanden-Eijnden.
\newblock Parameters as interacting particles: long time convergence and asymptotic error scaling of neural networks.
\newblock In {\em Advances in Neural Information Processing Systems}, pages 7146--7155, 2018.

\bibitem[Sam93]{samarov1993exploring}
Alexander~M Samarov.
\newblock Exploring regression structure using nonparametric functional estimation.
\newblock {\em Journal of the American Statistical Association}, 88(423):836--847, 1993.

\bibitem[Ser99]{servedio1999pac}
Rocco~A Servedio.
\newblock On {PAC} learning using {Winnow}, {Perceptron}, and a {Perceptron-like} algorithm.
\newblock In {\em Conference on Computational Learning Theory}, pages 296--307, 1999.

\bibitem[SS20]{sirignano2020mean}
Justin Sirignano and Konstantinos Spiliopoulos.
\newblock Mean field analysis of neural networks: A law of large numbers.
\newblock {\em SIAM Journal on Applied Mathematics}, 80(2):725--752, 2020.

\bibitem[Ste81]{stein1981estimation}
Charles~M Stein.
\newblock Estimation of the mean of a multivariate normal distribution.
\newblock {\em The Annals of Statistics}, pages 1135--1151, 1981.

\bibitem[Sto77]{stone1977consistent}
Charles~J Stone.
\newblock Consistent nonparametric regression.
\newblock {\em The Annals of Statistics}, pages 595--620, 1977.

\bibitem[Sto80]{stone1980optimal}
Charles~J Stone.
\newblock Optimal rates of convergence for nonparametric estimators.
\newblock {\em The Annals of Statistics}, pages 1348--1360, 1980.

\bibitem[TDD{\etalchar{+}}24]{troiani2024fundamentallimitsweaklearnability}
Emanuele Troiani, Yatin Dandi, Leonardo Defilippis, Lenka Zdeborová, Bruno Loureiro, and Florent Krzakala.
\newblock Fundamental limits of weak learnability in high-dimensional multi-index models.
\newblock {\em arXiv preprint arXiv:2405.15480}, 2024.

\bibitem[TWKS14]{trivedi2014consistent}
Shubhendu Trivedi, Jialei Wang, Samory Kpotufe, and Gregory Shakhnarovich.
\newblock A consistent estimator of the expected gradient outerproduct.
\newblock In {\em Thirtieth Conference on Uncertainty in Artificial Intelligence}, pages 819--828, 2014.

\bibitem[Vem10]{vempala2010learning}
Santosh~S Vempala.
\newblock Learning convex concepts from gaussian distributions with pca.
\newblock In {\em 51st Annual IEEE Symposium on Foundations of Computer Science}, pages 124--130, 2010.

\bibitem[Wat64]{watson1964smooth}
Geoffrey~S Watson.
\newblock Smooth regression analysis.
\newblock {\em Sankhy{\=a}: The Indian Journal of Statistics, Series A}, pages 359--372, 1964.

\bibitem[XTLZ02]{xia2002adaptive}
Yingcun Xia, Howell Tong, Wai~Keung Li, and Li-Xing Zhu.
\newblock An adaptive estimation of dimension reduction space.
\newblock {\em Journal of the Royal Statistical Society Series B: Statistical Methodology}, 64(3):363--410, 2002.

\bibitem[YBL17]{yang2017high}
Zhuoran Yang, Krishnakumar Balasubramanian, and Han Liu.
\newblock High-dimensional non-{Gaussian} single index models via thresholded score function estimation.
\newblock In {\em International Conference on Machine Learning}, 2017.

\bibitem[YXKH23]{yuan2023efficient}
Gan Yuan, Mingyue Xu, Samory Kpotufe, and Daniel Hsu.
\newblock Efficient estimation of the central mean subspace via smoothed gradient outer products.
\newblock {\em arXiv preprint arXiv:2312.15469}, 2023.

\end{thebibliography}

\end{document}